\documentclass[10pt]{article}
\usepackage{setspace}
\usepackage[margin=1.1in]{geometry}

\usepackage[font=footnotesize]{caption}

\usepackage{amsgen}
\usepackage{amsfonts}
\usepackage{amssymb}
\usepackage{amsbsy}
\usepackage{graphicx}
\usepackage[breaklinks=true,colorlinks=true,linkcolor=blue,urlcolor=blue,citecolor=blue]{hyperref}
\usepackage[ruled,vlined]{algorithm2e}
\usepackage{algorithmic}
\usepackage{color}
\usepackage{multirow}
\usepackage[xcolor=blue]{changes}
\usepackage{amsmath,amssymb}

\usepackage{caption}
\usepackage{subcaption}

\newcommand{\norm}[1]{\left\lVert#1\right\rVert}

\title{Domain Knowledge Driven 3D Dose Prediction Using Moment-Based Loss Function}
\author{Gourav~Jhanwar$^{1*}$,~Navdeep~Dahiya$^{2*}$,~Parmida~Ghahremani$^{1}$, ~Masoud~Zarepisheh$^{1\dagger}$,\\~Saad~Nadeem$^{1\dagger}$}
\date{}

\usepackage{fancyhdr}
\begin{document}
\maketitle

{\footnotesize
\noindent
$^1$Department of Medical Physics, Memorial Sloan-Kettering Cancer Center, New York, NY, USA.

\noindent
$^2$Department of Electrical \& Computer Engineering, Georgia Institute of Technology, Atlanta, GA, USA.

\noindent
$^*$ Co-first authors.

\noindent
$^\dagger$ Co-senior authors

\noindent
\textbf{Corresponding author:} Masoud Zarepisheh (zarepism@mskcc.org)
}

\begin{abstract}
\noindent

Dose volume histogram (DVH) metrics are widely accepted evaluation criteria in the clinic. However, incorporating these metrics into deep learning dose prediction models is challenging due to their non-convexity and non-differentiability. We propose a novel moment-based loss function for predicting 3D dose distribution for the challenging conventional lung intensity modulated radiation therapy (IMRT) plans. The moment-based loss function is convex and differentiable and can easily incorporate DVH metrics in any deep learning framework without computational overhead. The moments can also be customized to reflect the clinical priorities in 3D dose prediction. For instance, using high-order moments allows better prediction in high-dose areas for serial structures. We used a large dataset of 360 (240 for training, 50 for validation and 70 for testing)  conventional lung patients with 2Gy $\times$ 30 fractions to train the deep learning (DL) model using clinically treated plans at our institution. We trained a UNet like CNN architecture using computed tomography (CT), planning target volume (PTV) and organ-at-risk contours (OAR) as input to infer corresponding voxel-wise 3D dose distribution. We evaluated three different loss functions: (1) The popular Mean Absolute Error (MAE) Loss, (2) the recently developed MAE + DVH Loss, and (3) the proposed MAE + Moments Loss. The quality of the predictions was compared using different DVH metrics as well as dose-score and DVH-score, recently introduced by the \textit{AAPM knowledge-based planning grand challenge}. Model with (MAE + Moment) loss function outperformed the model with MAE loss by significantly improving the DVH-score (11\%, p$<$0.01) while having similar computational cost. It also outperformed the model trained with (MAE+DVH) by significantly improving the computational cost (48\%) and the DVH-score (8\%, p$<$0.01). The code, pretrained models, docker container, and Google Colab project along with a sample dataset are available on our DoseRTX GitHub (\url{https://github.com/nadeemlab/DoseRTX}).

\end{abstract}


\section{Introduction}
Despite recent advances in optimization and treatment planning, intensity modulated radiation therapy (IMRT) treatment planning remains a time-consuming and resource-demanding task with the plan quality heavily dependent on the planner's experience and expertise \cite{berry2016interobserver,das2009analysis, nelms2012variation}. This problem is even more pronounced for challenging clinical cases such as conventional lung with complex geometry and intense conflict between the objectives of irradiating planning target volume (PTV) and sparing organ at risk structures (OARs). Conventional lung plans are generally considered challenging in the clinic due to large PTV sizes, variation in the location of PTV, and having many sensitive nearby structures, making the planning process time-consuming and difficult. Balancing the trade-off between conflicting objectives can lead to sub-optimal plans~\cite{moore2015quantifying}, sacrificing the plan quality. 

In the last decade, several techniques have been developed to automate or facilitate the radiotherapy treatment planning process. Multi-criteria optimization (MCO)~\cite{craft2008many} facilitates the planning by generating a set of Pareto optimal plans upfront and allowing the user to navigate among them offline. Hierarchical constrained optimization enforces the critical clinical constraints using \textit{hard constraints} and improves the other desirable criteria as much as possible by sequentially optimizing these~\cite{zarepisheh2019automated, breedveld2009equivalence}. Knowledge-based planning (KBP) is a data-driven approach to automate the planning process by leveraging a database of pre-existing patients and learning a map between the patient anatomical features and some dose distribution characteristics. The earlier KBP methods used machine learning methods such as linear regression, principal component analysis, random forests, and neural networks to predict DVH as a main metric to characterize the dose distribution~\cite{DVH1,DVH2,DVH3,DVH4,DVH5,DVH6,DVH7}. However, DVH lacks any spatial information and only predicts dose for the delineated structures. 

More recently, deep learning (DL) methods have been successfully used in  radiation oncology for automated image contouring/segmentation \cite{han2017mr,ibragimov2017segmentation,men2017automatic} as well as 3D voxel-level dose prediction~\cite{TreatmentPlanningReview-Wang2020}. A typical DL dose prediction method uses a convolutional neural network (CNN) model which receives a 2D or 3D input in the form of planning CT with OAR/PTV masks and produces a voxel-level dose distribution as its output. The predicted dose is compared to the real dose using some form of loss function such as mean absolute error (MAE) or Mean square Error (MSE). The loss function in fact quantifies the goodness of the prediction by comparing that to the delivered dose voxel-by-voxel. While MAE and MSE are powerful and easy-to-use loss functions, they fail to integrate any domain specific knowledge about the quality of dose distribution including maximum/mean dose at each structure. The direct representation of DVH results in a discontinuous, non-differentiable, and non-convex function which makes it difficult to integrate it into any DL model. Nguyen et al.~\cite{DVHLoss} proposed a a continuous and differentiable, yet non-convex, DVH-based loss function (not to be confused with predicting DVH).  In this paper, we propose a differentiable and convex surrogate loss function for DVH using multiple moments of dose distribution. It has been previously shown that moments can approximate a DVH to an arbitrary accuracy~\cite{zinchenko2008controlling}, and also moments have been successfully used to replicate a desired DVH~\cite{zarepisheh2013moment}. The convexity and differentiability of the moment-based loss function makes it computationally appealing and also less prone to the local optimality. Furthermore, using different moments for different structures allows the DL model to drive the prediction according to the clinical priorities.

\section{Materials and Method}

\subsection{Loss Functions}

We use three types of loss functions in our study. First, we use mean absolute error (MAE) that measures the error between paired observations of real and predicted 3D dose. MAE is defined as $\frac{1}{N}\sum_i|D_p(i) - D_r(i)|$ where $N$ is the total number of voxels and $D_p$, $D_r$ are the predicted and real doses. We preferred to use MAE versus a common alternative, mean squared error (MSE), as MAE produces less blurring in the output compared to MSE~\cite{pix2pix2017}. {MAE loss is one of the widely used loss in machine learning and it is calculated as the mean of absolute difference between the predicted and reference dose. 
\begin{equation}
L_{MAE} = \frac{1}{n}\sum_{i=1}^{n} |d_{pred}^{i} - d_{ref}^{i}|
\end{equation}
where $d_{pred}^{i}$ is the predicted dose and $d_{ref}^{i}$ is the reference dose for the $i^{th}$ voxel.}

\subsubsection{Sigmoid-Based DVH Loss}
Nguyen et al. \cite{DVHLoss} proposed approximating the heaviside step function by the readily differentiable sigmoid function to address discontinuity and non-differentiability issues of DVH function. Borrowing the notation from~\cite{DVHLoss}, for a given volumetric dose distribution $D$ and a segmentation mask {$B_s$} for the $s^{th}$ structure, the volume-at-dose with respect to the dose $d_t$, denoted by $v_{s,t}(D,{B_s})$, is defined as the volume fraction of a given region-of-interest (OARs or PTV) which receives a dose of at least $d_t$ or higher which can be approximated as:
\begin{equation}\label{MAE-DVH-loss}
    v_{s,t}(D,{B_s}) = \frac{\sum_i \sigma\left( \frac{D(i) - d_t}{\beta} \right){B_s}(i)}{\sum_i{B_s}(i)}
\end{equation}
where $\sigma$ is the sigmoid function, $\sigma(x) = \frac{1}{1 + e^{-x}}$, $\beta$ is histogram bin width, and $i$ loops over the voxel indices of the dose distribution. {Based on this, the DVH for the structures $s$ is defined as,
\begin{equation}
    DVH(D,{B_s}) = (v_{s,d_1}, v_{s,d_2}, v_{s,d_3}, .., v_{s,{d_n}_t}).
\end{equation}}
The DVH loss can be calculated using MSE between the real and predicted dose DVH and is defined as follows:
\begin{equation}
    L_{DVH}\left( D_r, D_p, B\right) = \frac{1}{n_s} \frac{1}{n_t} \sum_s \norm{DVH\left(D_r,B_s\right) - DVH\left(D_p, B_s\right)}_2^2.
\end{equation}
{
Total loss for training the UNET using sigmoid based non-convex DVH loss is then given by,
\begin{equation}
    L_{Total} = L_{MAE} + w_{DVH}*L_{DVH}
\end{equation}
where $w_{DVH}$ is the weight for DVH loss. }
\subsubsection{Moment Loss}
Moment loss is based on the idea that a DVH can be well-approximated using a few moments \cite{zinchenko2008controlling, zarepisheh2013moment}:

$$ DVH \sim ( M_1, M_2, M_3, ..., M_{p} ) $$
where $M_{p}$ represents the moment of order $p$ defined as:
\begin{equation}
    M_{p} = \left(\frac{1}{|V_s|}\sum_{j\in V_s} d_j^p \right)^\frac{1}{p}
\end{equation}
where $V_s$ is a set of voxels belonging to the structure $s$, and $d$ is the dose. $M_1$ is simply the mean dose of a structure whereas $M_\infty$ represents the max dose, and for $p>1$, $M_{p}$ represents a value between mean and max doses.   
 
In our experiments, we used a combination of three moments $P = \{1, 2, 10\}$ for the critical OARs and PTV, where $M_{1}$ is exactly the mean dose, $M_{2}$ is the dose somewhere between the mean and max dose and $M_{10}$ approximates the max dose. 

The moment loss is calculated using mean square error between the actual and predicted moment for the structure: 
\begin{equation}\label{MAE-Moment-loss}
    L_{moment} = \sum_{p\in P} \norm{M_p - \bar{M}_p}_2^2
\end{equation}
where $M_p$ and $ \bar{M}_p$ are the $p^{th}$ moment of the actual dose and the predicted dose of a given structure, respectively.

{Total loss for training the neural network using moment based convex loss function is then defined as,
\begin{equation}
    L_{Total} = L_{MAE} + w_{Moment}*L_{Moment}
\end{equation}
where $w_{Moment}$ is the weight for the Moment loss.}

\subsection{Patient Dataset}
We used 360 randomly selected lung cancer patients treated with conventional IMRT with 60Gy in 30 fractions at Memorial Sloan Kettering Cancer Center between the year 2017 and 2020. All these patients received treatment and therefore included the treated plans which were manually generated by experienced planners using 5--7 coplanar beams and 6 MV energy. Table \ref{table:1} refers to the clinical criteria used at our institution. All these plans were generated using Eclipse$^{TM}$ V13.7-V15.5 (Varian Medical Systems, Palo Alto, CA, USA).

\begin{table}[ht!]
\caption{Clinical Max/Mean dose (in Gy) and Dose-volume criteria}
\label{table:1}
\centering
 \begin{tabular}
 {|l |c | c| c|} 
 \hline
 \textbf{Structure} & \textbf{Max (Gy)} & \textbf{Mean (Gy)} & \textbf{Dose-volume} \\ [0.5ex] 
 \hline
 PTV & 72 &  &  \\ 
 Lungs Not GTV	& 66 & 21 & V(20Gy) $<=$ 37\%\ \\
 Heart & 66 & 20 & V(30Gy) $<=$ 50\%\ \\
 Stomach & 54 & 30 & \\
 Esophagus & 66 & 34 & \\
 Liver & 66 &  & V(30Gy) $<=$ 50\%\ \\ 
 Cord & 50 & & \\
Brachial Plexus	& 65 &  & \\ 
 \hline
\end{tabular}
\end{table}

\subsection{Inputs and Preprocessing}
\begin{figure}[htb!]
    \centering
    \includegraphics[width=0.98\linewidth]{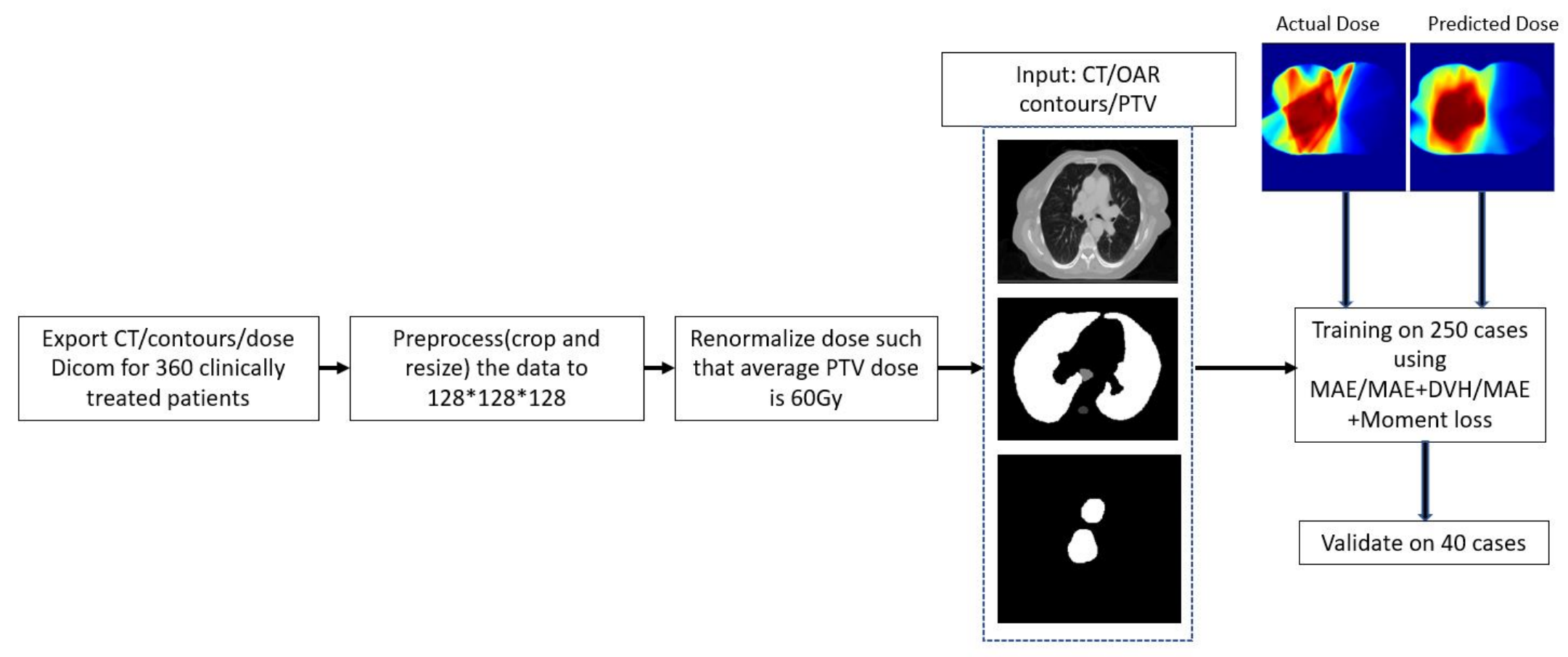}%
    \caption{Overview of data processing pipeline and training a 3D network to generate a 3D voxelwise dose. OARs are one-hot encoded and concatenated along the channel axis with CT and PTV input to the network.}
    \label{fig:pipeline_nobeam}
\end{figure}

Structure contours and the 3D dose distribution were extracted from the Eclipse V15.5 (Varian Medical Systems, Palo Alto, CA, USA). Each patient has a planning CT and manually delineated contours of PTV and OARs which may differ from patient to patient depending on the location and size of the tumor. However, all patients have esophagus, spinal cord, heart, left lung, right lung and PTV delineated. Hence, we use these five OARs and PTV as inputs in addition to the planning CT. Figure~\ref{fig:pipeline_nobeam} shows the overall workflow to train a CNN to generate voxel-wise dose distribution.
The CT images may have different spatial resolutions but have the same in-plane matrix dimensions of 512$\times$512. The PTV and OAR segmentation dimensions match those of the corresponding planning CTs. The intensity values of the input CT images are first clipped to have range of [-1024, 3071] and then re-scaled to range [0, 1] for input to the DL network. The OAR segmentations are converted to a one-hot encoding scheme with value of 1 inside each anatomy and 0 outside. The PTV segmentation is then added as an extra channel to the one-hot encoded OAR segmentation.

The dose data have different resolutions than the corresponding CT images. Each pair of the doses is first re-sampled to match the corresponding CT image.  For easier training and comparison between different patients, the mean dose inside PTV of all patients is re-scaled to 60 Gy. This serves as a normalization for comparison between patients and can be easily shifted to a different prescription dose by a simple re-scaling inside the PTV region. 

Finally, in order to account for the GPU RAM budget, we crop a 300$\times$300$\times$128 region from all the input matrices (CT/OAR/PTV/Dose) and re-sample it to a consistent 128$\times$128$\times$128 dimensions. We used the OAR/PTV segmentation masks to guide the cropping to avoid removing any critical regions of interest.

\subsection{CNN Architecture}
Unet is a fully connected network which has been widely used in the medical image segmentation. We train a Unet like CNN architecture~\cite{ronneberger2015unet,pix2pix2017} to output the voxel-wise 3D dose prediction corresponding to an input comprising of 3D CT/contours which are concatenated along the channel dimension. The network follows a common encoder-decoder style architecture which is composed of a series of layers which progressively downsample the input (encoder) using max pooling operation, until a bottleneck layer, where the process is reversed (decoder). Additionally, Unet-like skip connections are added between corresponding layers of encoder and decoder. This is done to share low-level information between the encoder and decoder counterparts.

\begin{figure}[htb!]
    \centering
    \includegraphics[width=\linewidth]{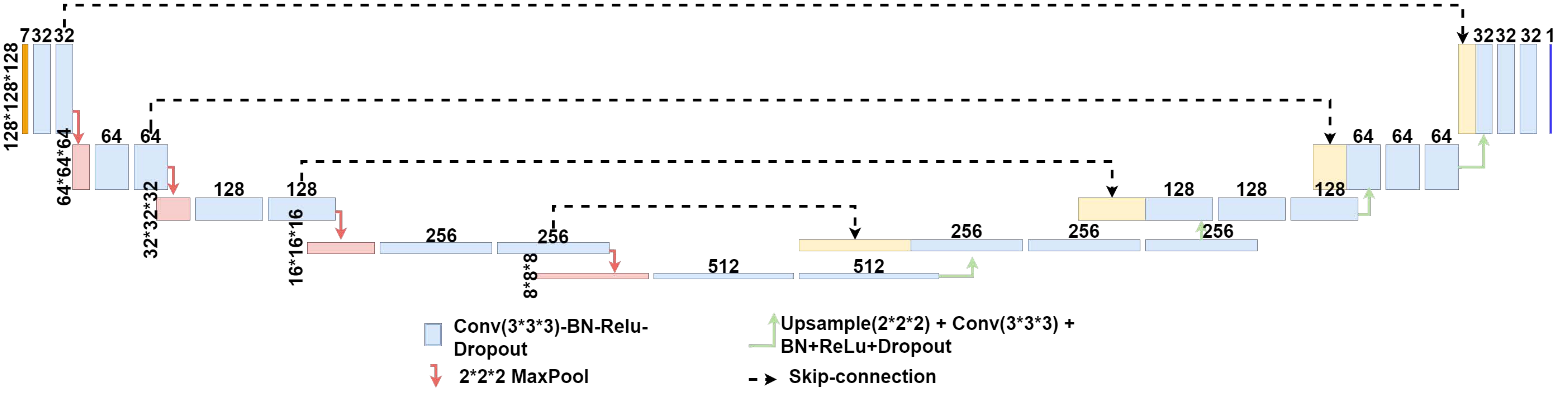}%
    \caption{3D Unet architecture used to predict 3D voxelwise dose.}
    \label{fig:network_arch}
\end{figure}

The network (Figure~\ref{fig:network_arch}) uses Convolution-BatchNorm-ReLU-Dropout as a block to perform series of convolution. Dropout is used with a dropout rate of $50\%$. Maxpool is used to downsample the image by 2 in each spatial level of encoder. All the convolutions in the encoder are $3\times3\times3$ 3D spatial filters with a stride of 1 in all 3 directions. In the decoder we use trilinear upsampling followed by regular $2\times2\times2$ stride 1 convolution. The last layer in the decoder maps its input to a one channel output ($128^3, 1$).

\subsection{Evaluation Criteria}
To evaluate the quality of the predicted doses, we adopt the metrics used in recent AAPM ``open-access knowledge-based planning grand challenge'' (OpenKBP) \cite{babier2021openkbp}. This competition was designed to advance fair and consistent comparisons of dose prediction methods for knowledge-based planning in radiation therapy research. The competition organizers used two separate scores to evaluate dose prediction models: dose score, which evaluates the overall 3D dose distribution and a DVH score, which evaluates a set of DVH metrics. The dose score was simply the MAE between real dose and predicted dose. The DVH score which was chosen as a radiation therapy specific clinical measure of prediction quality involved a set of DVH criteria for each OAR and target PTV. Mean/D(0.1cc) dose received by OAR was used as the DVH criteria for OAR while PTV had three criteria: {D99, D95 and D1 which are the doses received by 99\% ($1^{st}$ percentile), 95\% ($5^{th}$ percentile), and 1\% ($99^{th}$ percentile) of voxels in the target PTV}. DVH error, the absolute difference between the DVH criteria for real and predicted dose, was used to evaluate the DVHs. Average of all DVH errors was taken to encapsulate the different DVH criteria into a single score {called "DVH Score"}, measuring the DVH quality of the predicted dose distributions. {We also report clinical evaluation criteria such as homogeneity index(HI) \cite{hodapp2012icru} given by $(\frac{D2-D98}{D50})$ and paddick conformity index(PCI) \cite{paddick2000simple} which is given by, $PCI = \frac{TV_{PIV^2}}{TV*PIV}$, where $TV$ = Target volume, $PIV$ = Prescription Isodose volume and $TV_{PIV}$ = volume of the target covered by the prescription isodose.}

\subsection{Deep Learning Settings}
In our experiments, we used Stochastic Gradient Descent (SGD), with a batch size of 1, and Adam optimizer~\cite{DBLP:journals/corr/KingmaB14} with an initial learning rate of $0.0002$, and momentum parameters $\beta_1 = 0.5$, $\beta_2 = 0.999$. We trained the network for total of 200 epochs. We used a constant learning rate of $0.0002$ for the first 100 epochs and then let the learning rate linearly decay to 0 for the final 100 epochs. When using the MAE and DVH combined loss, we scaled the DVH component of the loss by a factor of 10. Also, when using the MAE and Moment combined loss, we used the weight of 0.01 for moment loss based upon our validation results.

We divided our training set of 290 images into train/validation set of 240 and 50 images respectively and determined the best learning rate and scaling factor for (MAE + DVH) loss and (MAE + Moment) loss. Afterwards, we trained all our models using all 290 training datasets and tested on the holdout 70 datasets used for reporting results.

We created the implementations of the CNN model, loss functions and other related training/testing scripts in pytorch and we conducted all our experiments on an NVIDIA A40 GPU with 48 GB VRAM.

\section{Results}
\begin{figure*}[t!]
\begin{center}
\footnotesize
\setlength{\tabcolsep}{8pt}
\begin{tabular}{cc}
\includegraphics[width=0.46\textwidth]{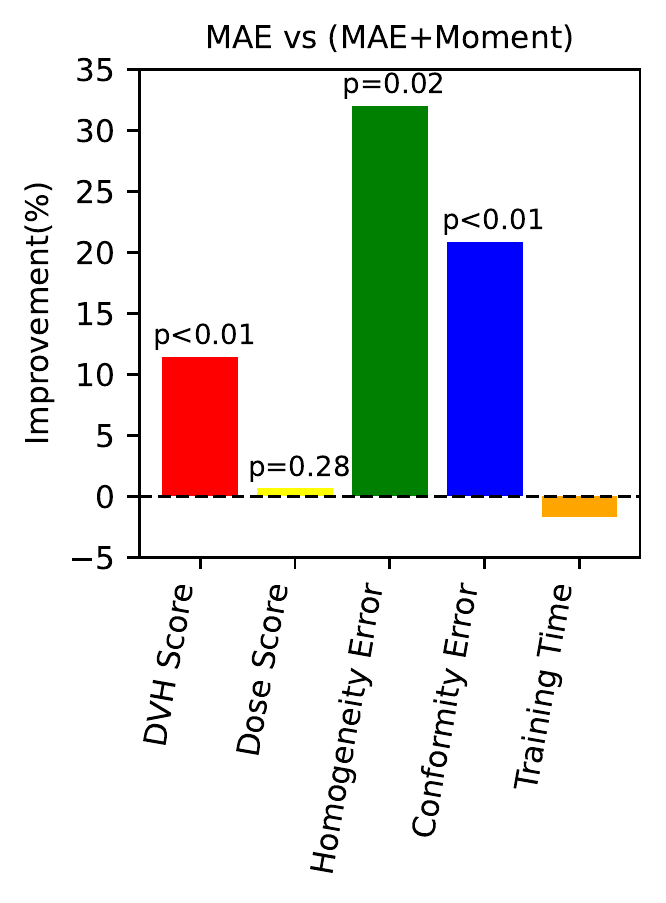}&
\includegraphics[width=0.46\textwidth]{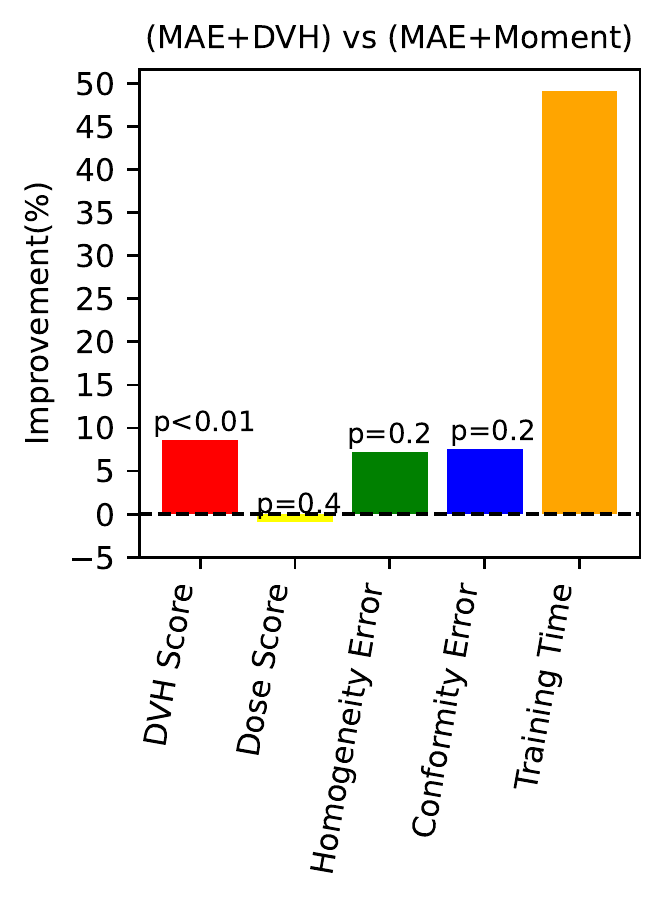}\\
(a) & (b)\\
\end{tabular}
\end{center}
\caption{Comparison of different metrics for (a) MAE vs (MAE + Moment) and (b) (MAE+ DVH) vs (MAE + Moment) losses. Y axis shows the relative improvement(in \%) using (MAE+Moment) loss compared to MAE and (MAE+DVH) loss. The higher is always better. For statistical analysis, Wilcoxon signed-rank test was used and p = 0.05 was considered statistically significant. }
\label{fig:metrics}
\end{figure*}

{Figure~\ref{fig:metrics}(a) compares the model trained using MAE loss vs (MAE+Moment) loss with respect to different metrics. The y-axis represents the relative improvement obtained using (MAE+Moment) loss, comparing the prediction and ground truth (manual plan).   While both models performed similarly with respect to the dose-score and training time ($\sim$ 7 hrs), the (MAE+Moment) loss improved the DVH score, homogeneity error and conformity error by 11$\%$, 32$\%$, and 21$\%$ respectively. For statistical tests, Wilcoxon signed-rank test was performed for different metrics and p = 0.05 was considered as statistical significance. The difference in DVH-score and conformity error was found  statistically significant(p $<$ 0.01).}

{Similarly, Figure~\ref{fig:metrics}(b) compares the model trained using (MAE+DVH) loss vs (MAE+Moment) loss and shows the relative improvement obtained by (MAE+Moment) loss. (MAE + Moment) significantly improves the training time (almost half), while modestly improves the DVH score, homogeneity  and conformity errors (about 7$\%$-8$\%$). The significant improvement in training time for (MAE+Moment) loss  owes to its convexity and simplicity. Statistically significant difference(p $<$ 0.01) was observed for DVH score.}

\begin{figure}[htb!]
    \centering
    \includegraphics[width=0.95\linewidth]{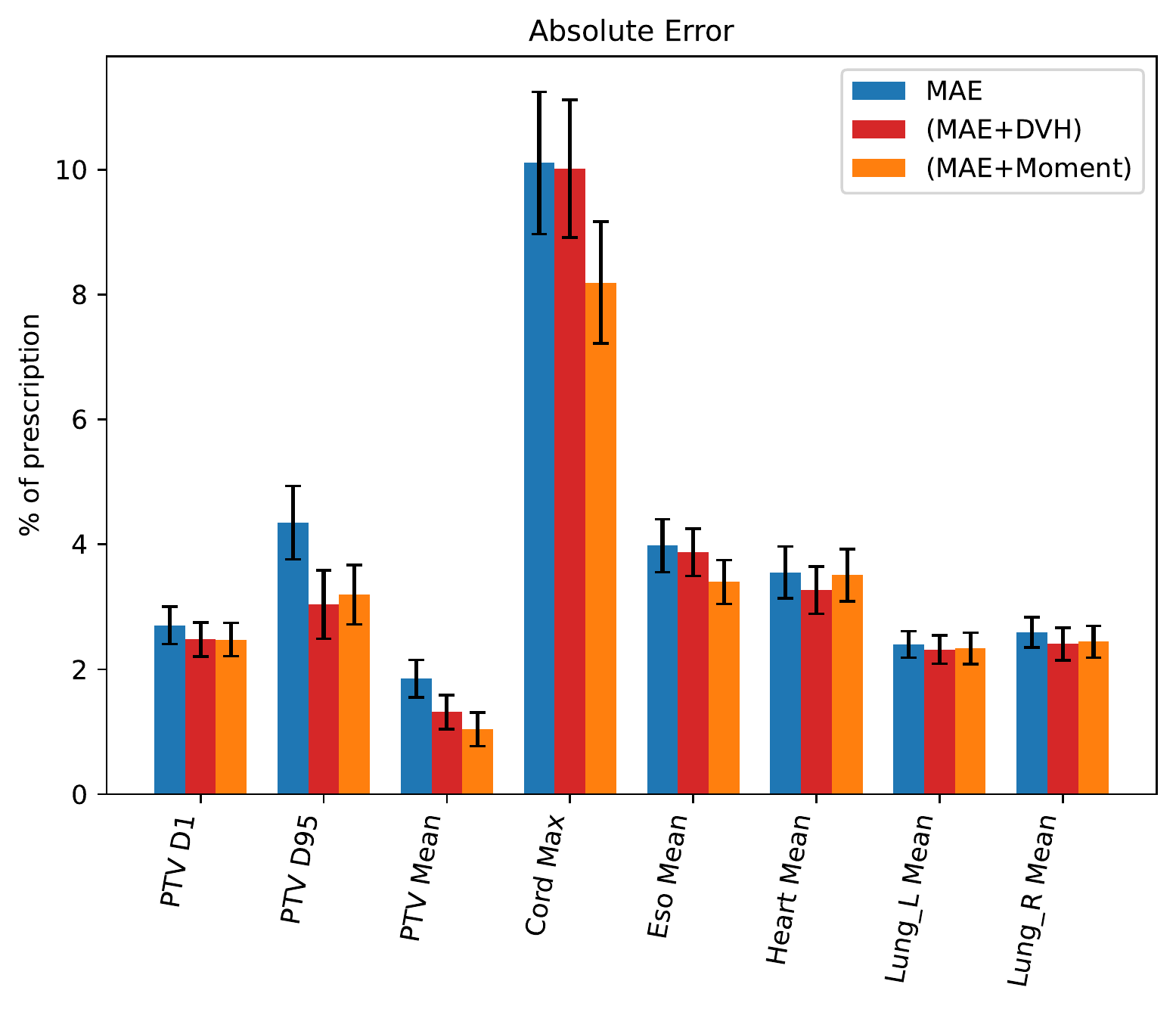}%
    \caption{Absolute Error (in percentage of prescription) for clinically relevant max/mean dose for organ-at-risk and PTV D1/D95 and mean dose. The lower is always the better. Error bars represents 95\% confidence interval ($\bar{x} \pm 1.96*\frac{\sigma}{\sqrt{n}}$) for all the test patients.}
    \label{fig:AbsoluteError}
\end{figure}

Figure~\ref{fig:AbsoluteError} shows the average absolute error between actual and predicted dose in terms of percentage of prescription for different clinically relevant criteria. Critical OARs like cord and esophagus showed substantial improvement in max and mean absolute dose error respectively using (MAE + Moment) loss compared to other two in the category. {PTV D95 and mean dose} showed marginal improvements in the dose prediction quality compared to MAE loss. There was small/no-improvement in the mean absolute error for other healthy organs (i.e., left lung, right lung, heart).

\begin{figure}[htb!]
    \centering
    \includegraphics[width=0.95\linewidth]{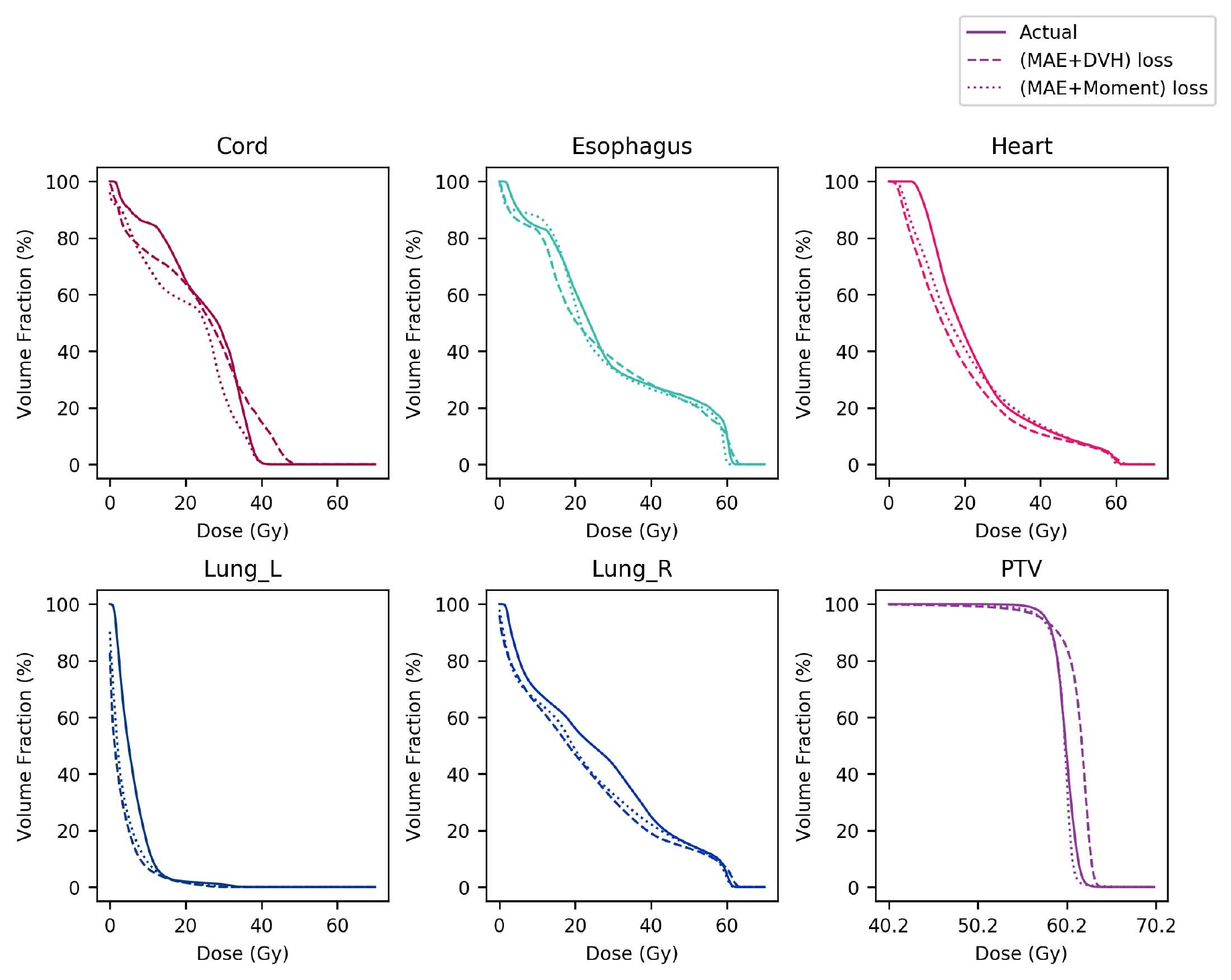}%
    \caption{DVH plots for different structures using i) Actual Dose, predicted dose using ii) MAE+DVH loss, and iii) MAE+Moment loss, for one patient.}
    \label{fig:DVHPlots}
\end{figure}

Figure~\ref{fig:DVHPlots} compares the DVH of an actual dose (ground-truth here) with {two} predictions obtained from two different loss functions for a patient. As can been seen, in general, the prediction generated with (MAE + Moment) loss resembles the actual ground-truth dose more than the other model, for this particular patient, especially for PTV.

\begin{figure}[htb!]
    \centering
    \includegraphics[width=0.95\linewidth]{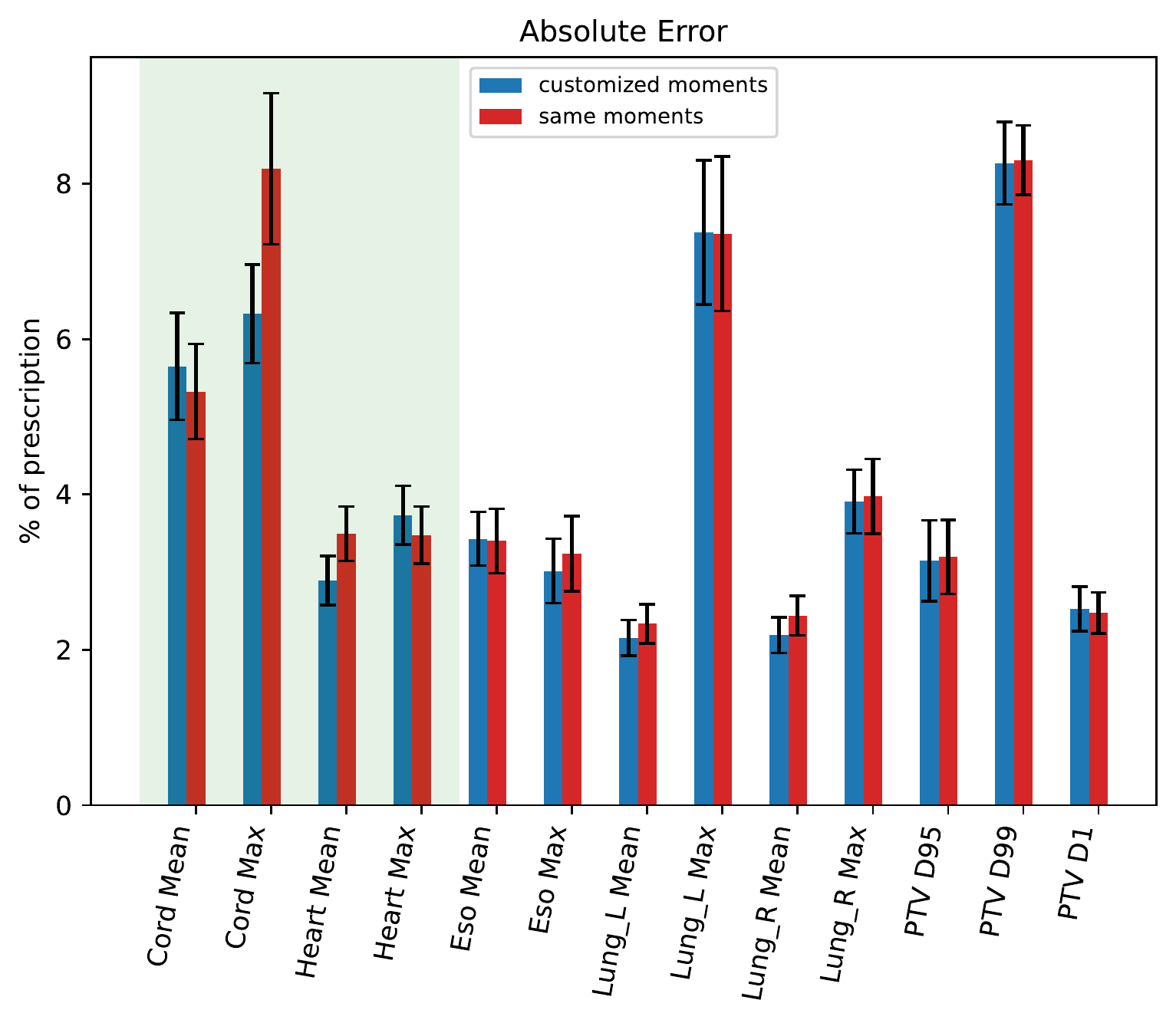}%
    \caption{Absolute Error (in percentage of prescription) for max/mean dose for organ-at-risk and PTV D1/D95/D99 dose. The lower is always the better. Absolute error in the green region shows the impact of customized moments. Error bars represents 95\% confidence interval ($\bar{x} \pm 1.96*\frac{\sigma}{\sqrt{n}}$) for all the test patients.}
    \label{fig:SensitivityAnalysis1}
\end{figure}

Figure~\ref{fig:SensitivityAnalysis1} shows the comparison of the absolute error for the model trained with default moments ($p=1,2,10$) for all the structures (red bar) and the model that used different moments for cord ($p=5,10$) and heart ($p=1,2$) and default moments ($p=1,2,10$) for all other structures (blue bar). As can be seen in the figure, using higher-order moments for cord improves the maximum dose prediction, while using lower-order moments for heart improves the mean dose prediction. 

\section{Discussion}
In this study, we have employed moments as a surrogate loss function to integrate DVH into deep learning (DL) 3D dose prediction. Moments provide a mathematically rigorous and computationally efficient way to incorporate DVH information in any DL architecture without any computational overhead. This allows us to incorporate the domain specific knowledge and clinical priorities into the DL model. Using MAE + Moment loss means the DL model tries to match the actual dose (ground-truth) not only at a micro-level (voxel-by-voxel using MAE loss) but also at a macro-level (structure-by-structure using representative moments).

Moments are essentially simple polynomial functions which can be calculated efficiently. Given their convexity, they do not suffer from the local optimality issue, making them  more reliable choices with more robust behaviour against the stochastic nature of the optimization techniques, commonly used in deep learning models. The computational efficiency of the moments allows training of large DL models. They also offer better fine-tuning of the \textit{hyper parameters}.  

The moments in conjunction with MAE help to incorporate DVH information into the DL model, however, the MAE loss still plays the central role in the prediction. In particular, the moments lack any spatial information about the dose distribution which is provided by the MAE loss. The MAE loss has also been successfully used across many applications and its performance is  well-understood. Further research is needed to investigate the performance of the moment loss on more data especially with different disease sites.  

The 3D dose prediction can facilitate and accelerate the treatment planning process by providing a reference plan which can be fed into a treatment planning optimization framework to be converted into a deliverable Pareto optimal plan. The \textit{dose-mimicking} approach has been commonly used in the literature \cite{fan2019automatic}, seeking the closest deliverable plan to the reference plan using quadratic function as a measure of distance. Babier et al. \cite{babier2020importance} proposed an inverse optimization framework which estimates the objective weights from the reference plan and then generates the deliverable plan by solving the corresponding optimization problem. Any improvements in the prediction, including the ones we have obtained using our proposed moment loss, needs to be ultimately evaluated using the entire pipeline of predicting a plan and converting that into a deliverable plan. 

\section{Conclusion}
This work shows that the moments are powerful tools with sound mathematical properties to integrate DVH as an important domain knowledge into 3D dose prediction without any computational overhead. The idea has been validated on a large dataset of 360  conventional lung patients. The conventional lung patients are usually considered challenging cases in the clinic due to their large PTV sizes and sensitive nearby critical structures (e.g., esophagus, heart).

\section*{Acknowledgments}
This project was partially supported by MSK Cancer Center Support Grant/Core Grant (P30 CA008748).

\end{document}